\documentclass[fleqn,10pt]{article}
\usepackage{microtype}
\usepackage{setspace}
\usepackage{hyperref} 
\usepackage{arxiv}
\usepackage[utf8]{inputenc} 
\usepackage[T1]{fontenc}    
\usepackage{url}            
\usepackage{booktabs}       
\usepackage{amsfonts}       
\usepackage{nicefrac}       
\usepackage{microtype}      
\usepackage{lipsum}         
\usepackage{graphicx}
\usepackage{cite}
\usepackage{doi}
\usepackage{amsmath,amssymb,amsfonts}
\usepackage{graphicx}
\usepackage{textcomp}
\usepackage{xcolor}
\usepackage{booktabs} 
\usepackage{graphicx}
\usepackage{epstopdf}
\usepackage{url}
\usepackage{multirow}
\usepackage{caption}
\usepackage{amssymb}
\usepackage{amsfonts}
\usepackage{float}
\usepackage{bm}
\usepackage{makecell}
\usepackage{dsfont}
\usepackage[T1]{fontenc}
\usepackage[left,modulo]{lineno}
\usepackage[ruled, vlined, linesnumbered, longend]{algorithm2e}
\usepackage{psfrag}
\usepackage{subfigure}
\usepackage{fancyhdr}
\usepackage{eucal}
\usepackage[english]{babel}
\usepackage{ifpdf}
\usepackage{cleveref} 
\usepackage{textcomp}
\usepackage{tikz}
\usepackage{tikz-inet}
\usetikzlibrary{calc,shapes.geometric}
\usetikzlibrary{shapes,snakes,arrows,automata} 
\usetikzlibrary{patterns}
\usepackage{multirow}
\usepackage{multicol}
\usepackage{authblk}

\newcommand{\espacio}[1]{}

\begin{document}
\include{macros}
\include{macrosBaster}
%
%
%
\renewcommand\Authfont{\bfseries}
\setlength{\affilsep}{0em}

\title{Unsupervised Assessment of Landscape Shifts Based on Persistent Entropy and Topological Preservation\thanks{Accepted article in KDD'2024, in the workshop on Drift Detection and Landscape Shifts. Slides of the presentation are available in  \href{https://aiimlab.org/events/KDD_2024_Discovering_Drift_Phenomena_in_Evolving_Landscape.html}{KDD-Delta}.}}
%

%
\newbox{\orcid}\sbox{\orcid}{\includegraphics[scale=0.06]{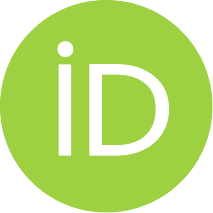}} 
\author{\href{https://orcid.org/0000-0002-9172-0155}{\usebox{\orcid}\hspace{1mm}Sebasti\'an Basterrech}}%
\affil{Department of Applied Mathematics and Computer Science, Technical University of Denmark, Denmark\\
{\{\texttt{sebbas@dtu.dk}\}}
}
\date{}
\renewcommand{\shorttitle}{Unsupervised Assessment of Landscape Shifts Based on Persistent Entropy and Topological Preservation}
\maketitle
\begin{abstract}
In Continual Learning (CL) contexts, concept drift typically refers to the analysis of changes in data distribution. A drift in the input data can have negative consequences on a learning predictor and the system's stability. The majority of concept drift methods emphasize the analysis of statistical changes in non-stationary data over time. In this context, we consider another perspective, where the concept drift also integrates substantial changes in the topological characteristics of the data stream.
In this article, we introduce a novel framework for monitoring changes in multi-dimensional data streams. 
%
We explore variations in the topological structures of the data, presenting another angle on the standard concept drift.
Our developed approach is based on persistent entropy and topology-preserving projections in a CL scenario. The framework operates in both unsupervised and supervised environments. To show the utility of the proposed framework, we analyze the model across three scenarios using data streams generated with MNIST samples. The obtained results reveal the potential of applying topological data analysis for shift detection and encourage further research in this area.
\end{abstract}
%
\onehalfspacing
\section{Introduction}
In continual learning scenarios, designing a machine learning (ML) model that is robust to distribution shifts is a crucial objective.
Traditional ML methods are susceptible to data perturbations, and shifts in input data distribution can significantly affect the model's performance.
Concept drift detectors encompass a family of techniques developed to analyze and detect distribution changes in the context of streaming data and time series.
The concept is based on changes in the statistical characteristics of the data over time~\cite{gama2014survey}.
However, there are scenarios where certain modifications in the distributions are not relevant. For example, simple translations or scaling of the data may not provide meaningful information in some contexts.
Therefore, it would be beneficial to define concept drift detectors that can detect distribution changes regardless of certain distortions or rotations of the data.
%
%
There are objects that are essentially equivalent to each other if we consider ``equivalence'' in the sense that it is possible to define a simple continuous transformation that approximates one object to the other. 
The essence of an object remains unchanged under simple transformations, such as rotation, translation, scaling, and other types of continuous transformations~\cite{Dai2021}.
On the other hand, there are objects that are essentially different, as it is not feasible to find any  continuous transformation to transform one object into another~\cite{Carlsson2009}. Or at least, it is not easy to find such a transformation with low computational resources.
The field of Topological Data Analysis (TDA), specifically through the mathematical formalism of algebraic topology, defines these concepts of equivalences and differences between objects. 
Persistent Entropy (PE) is a measure based in Shannon entropy that provides a summary of the geometric information derived from the topological features of a cloud of points~\cite{Atienza2019,Myers2019}.
It has been successfully used to effectively distinguish chaotic and periodic time series~\cite{Myers2019}.

\begin{figure}
\begin{minipage}[t]{1\textwidth}
\hfill
\begin{center}
\scalebox{0.8}{
\tikz{
%
\node (img1) at (0,0) {
{\includegraphics[scale=0.2]{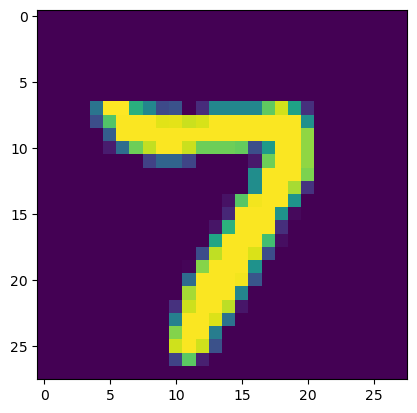}};
};
\node (img2) at (5,0) {\includegraphics[scale=0.2]{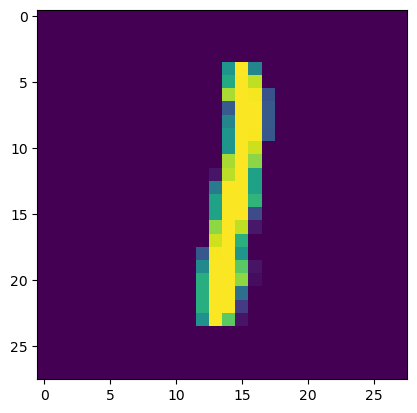}};
\node (img3) at (10,0) {\includegraphics[scale=0.2]{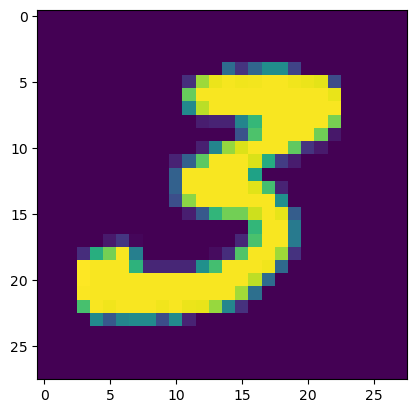}};
\draw[->,black,thick] (img1) -- (img2);
\node (text1) at (2.5,-0.5) {Stretching};
\draw[->,black,thick] (img2) -- (img3);
\node (text1) at (7.5,-0.5) {Bending};
\node (img3) at (0,-2.5) {
{\includegraphics[scale=0.2]{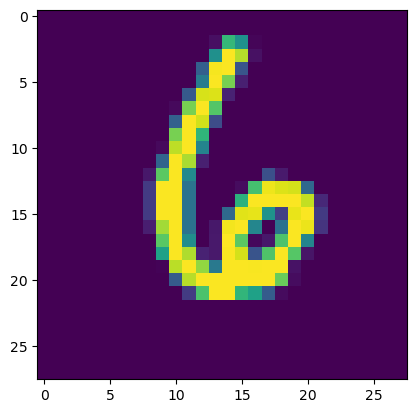}};
};
\node (img4) at (5,-2.5) {\includegraphics[scale=0.2]{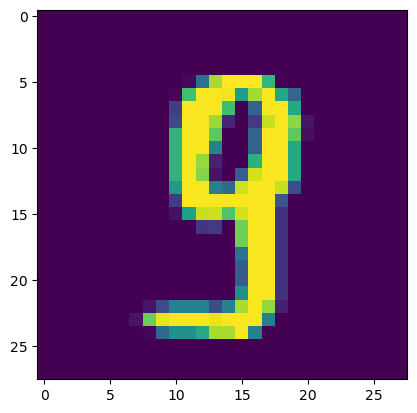}};
\node (img5) at (10,-2.5) {\includegraphics[scale=0.2]{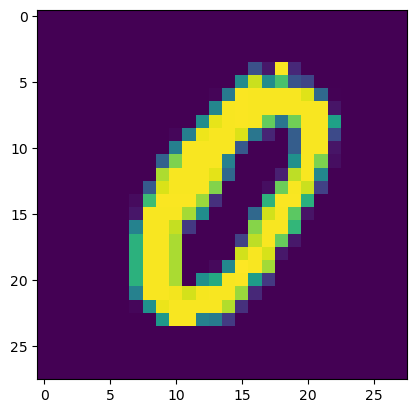}};
\draw[->,black,thick] (img3) -- (img4);
\node (text1) at (2.5,-3) {Rotation};
\node (text1) at (2.5,-3.7) {Bending};
\draw[->,black,thick] (img4) -- (img5);
\node (text1) at (7.5,-3) {Molding};
}}
\end{center}
\end{minipage}
\caption{Examples of objects that can be deformed and transformed into another object, equivalent in terms of topology. However, some shapes cannot be considered equivalent because it is not possible to define a sequence of simple continuous transformations to deform an original shape into another one while maintaining the original structure. For instance, from any of the digits at the top of the figure, we cannot form any of the digits at the bottom of the figure.}
\label{Transformations}
\end{figure}
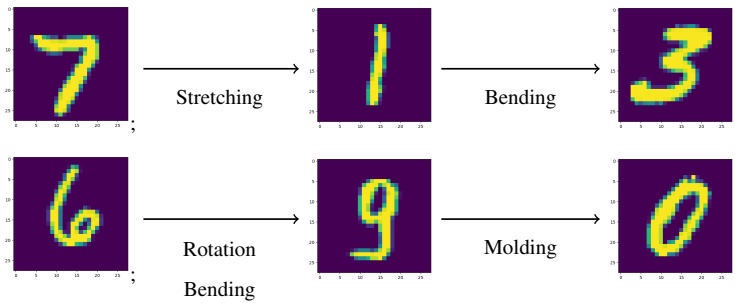

In this work, we introduce an extension of the classic concept drift that integrates algebraic topology.
We extend the concept of drift, which is based on statistical and geometrical information, with another concept that incorporates the notion of \textit{essential sameness} and \textit{essential difference}~\cite{Carlsson2009}.
Our work aims to provide insight into the research question: Is it adequate to identify a drift between two sequences if there exists a simple continuous bijective function that transforms one sequence into the other? In other words, is it adequate to identify a drift between two sequences when they are equivalent objects in terms of topology?
%
We observe a drift when the geometric characteristics of a point cloud \textit{essentially} change, becoming different from those of another point cloud.
Figure~\ref{Transformations} illustrates examples of objects that can be deformed using simple continuous transformations such as rotation, stretching, bending, and scaling. 
The first row of the figure displays three digits that are topologically equivalent.
The second row also shows three topologically equivalent objects.
It is not possible to transform any object from the first row into an object of the second row.
To quantify these types of topological variations, we take into account concepts provided by TDA.
We empirically investigate the changes in the distribution of high-dimensional data in an unsupervised scenario, using tools from persistence homology.
We develop a general-purpose framework that projects the input data into a low-dimensional space using a dimensionality reduction technique that preserves the topological features of the data. We then apply metrics of persistent homology to evaluate significant changes.
The projection is made using Self-Organizing Maps (SOM), which is a mapping technique to reduce dimensionality while preserving the topological characteristics of the input space~\cite{Kohonen2013}.
In the latent space, we explore the potential of persistent homology to find significant differences among data coming from consecutive chunks. 
We use the metric of persistent entropy, which summarizes the analysis of persistent homology in a single value~\cite{Atienza2020}.
In summary, this work offers the following contributions. 
\begin{itemize}
\item[(i)] We introduce a general-purpose framework for concept drift detection that operates in both supervised and unsupervised environments. 
This framework utilizes dimensionality reduction through a topological preserving mapping and evaluates significant changes using persistent homology.
\item[(ii)] The framework delivers results using a p-value score. When each chunk of data arrives, a non-parametric statistical test is performed, facilitating easy monitoring of drifts.
The hypothesis test is conducted on the values of persistent entropy. Since the framework provides a p-value score, the decision regarding the absence or presence of a drift is both robust and fast.
\item[(iii)] We provide an initial experimental evaluation with promising results across three case studies. We compare three dimensionality reduction techniques to evaluate the impact of preserving topological features when projecting data from the input space to the latent space. The results show  the benefits of combining a topology-preserving mapping with information regarding persistent homology.
\end{itemize}

\section{Background}
%
Drift detection techniques constitute a family of computational methods for detecting changes in the distribution of time series and data streams.
Shifts in data distribution can occur in different forms, probably the most accepted categories are: sudden, gradual, and incremental shifts~\cite{SouzaChallenges2020,Wozniak2023}.
A majority of drift detection techniques employ a classifier to categorize incoming instances, and the predictor generates a class label for each input instance, which is then compared to the actual class label.
Subsequently, the accuracy is assessed and utilized as a tool to determine whether a drift has occurred.
When the classifier's accuracy significantly decreases, then it is assumed that the data distribution has changed~\cite{Gama:2004,Goncalves2014,Goldenberg:2019}.
In this scenario, the effectiveness of various ensemble classifiers has been examined~\cite{du2014selective,Kolter:2003,Lapinski:2018,Maciel:2015}.
However, this approach can be applied only in a supervised context, and it requires the presence of ground truth labels, which are not always available. 
Another set of methods relies on empirically estimated distributions and statistical tests directly over the raw data~\cite{Blanco:2015,Clemmensen2023data,Hinder2020,Sobolewski:2013}.
These approaches may be sensitive to outliers and noise, and raw data analysis (e.g., applying density estimation) may also be affected by the curse of dimensionality~\cite{Basterrech2023ISDA,Basterrech2022SMC}.
Several studies propose to compare summary of statistics and aggregation metrics of the raw data, for instance, Cumulative Sum and Exponentially Weighted Moving Average~\cite{Ross:2012}. For a more comprehensive review of the latest advancements in the use of data descriptors for concept drift detection, see~\cite{Faber21,Hinder2022,Rabanser2019}.

Persistent homology is a key instrument in TDA as it may be used to describe the inherent structure of complex objects such as manifolds~\cite{Rieck2018}.
Specifically, persistent homology studies the evolution of $k$-dimensional topological features (often referred to as \textit{holes}) along a sequence of high-dimensional complex objects (named \textit{simplicial complexes})~\cite{Atienza2019,Carlsson2009}.
We understand topological features as shapes or data that remain unchanged under certain continuous transformations, such as connected components, independent cycles, and holes~\cite{Carlsson2009}.
This process can be seen as tracking changes across filtrations at multiple scales, following a specific algorithm that analyzes the connectivity information among the data points~\cite{Carlsson2009}.
Persistent Entropy, based in Shannon entropy,  provides a summary of the information derived from persistent homology~\cite{Atienza2019}. 
It is a measure for finding significant differences in the geometrical distribution of data points~\cite{Atienza2019,Oner2023}.
For a comprehensive and detailed exploration about {TDA} and persistent homology, see~\cite{Carlsson2009,Oner2023}.

%
%
%
%

\section{Methodology}
\label{Methodology}
This section outlines the contributions made in this brief article. First, we discuss the approach for transforming the input patterns into a different landscape that simplifies the analysis of drifts. Next, we introduce the process for estimating geometric changes between data points in different chunks. Finally, we present the sequence of modules that compose the developed framework.

\subsection{Creation of the latent space}
Monitoring and detecting distribution shifts is specially harder in the case of high-dimensional data. 
Although some attempts have been introduced in the literature for sparse multivariate time series~\cite{Shang2022,Zhang2019}, the scalability of existing algorithms remains an issue. In particular, methods based on probability mass distribution encounter significant challenges in high-dimensional spaces.
In addition, the computation of distances between vectors also has limitations in a high-dimensional space (e.g. Euclidean norm)~\cite{Sobczyk2022}, and a similarity analysis in the original input space may be computationally expensive.
Consequently, it is often more resource-efficient to first convert the data into a latent space, and then carry out the similarity analysis.
For this reason, a common approach is to transform the input data into a latent space, instead of making the analysis directly in the original space.
We analyze an approach that projects the input points into a latent space using dimensionality reduction (DR) techniques, which is a common method for handling data in high dimensions.
Here, we investigate the projections generated by {Self Organizing Maps} (SOMs) (also called Kohonen networks), and we compare the results with other two popular DR models (a linear projection (PCA) and the Kernel PCA).
The selection of an adequate data descriptor is crucial for ensuring a proper geometry in the latent space preserving the main features of the original space.

SOM is a bio-inspired method that combines concepts from Hebbian learning, vector quantization, and competitive learning~\cite{Fyfe05,Kohonen2013}.
Real-world data most often contain redundancies and inherent correlations among the variables.
SOM is a two-layered neural network that transforms intricate relationships among high-dimensional data into straightforward geometric relationships on a standard lattice, typically a two-dimensional grid~\cite{Ferles2018}.
Despite its simplicity, the SOM model is effective as a DR method, a clustering method, and a visualization tool for high-dimensional data~\cite{Saraswati2018}.
Another advantage is that the method is applicable to unsupervised problems and has the capability to preserve the most important topological features of the reference data~\cite{Kohonen2013}.

\subsection{Assessing shifts in the latent space}
\begin{figure}[htbp]
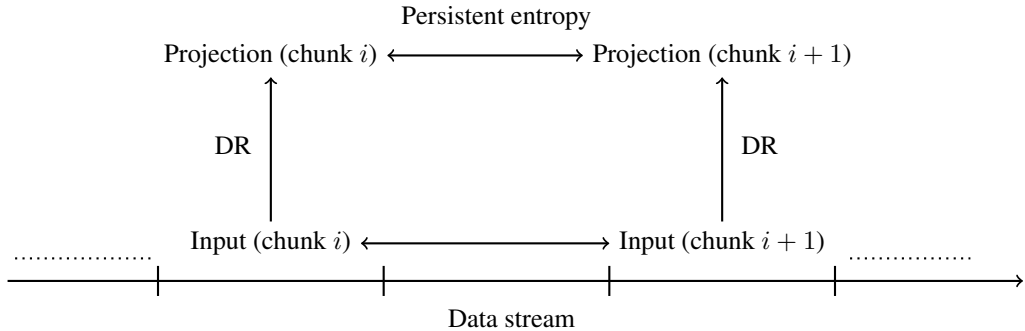

\centering
\scalebox{1}{
\tikz{
\node[rotate around={0:(0,0)},thick, minimum size=3.5ex]  (a0) at (4.5,3.5) {Persistent entropy};
\node[rotate around={0:(0,0)},thick, minimum size=3.5ex]  (m0) at (1.5,3) {Projection (chunk $i$)};
\node[rotate around={0:(0,0)},thick, minimum size=3.5ex]  (m1) at (7.5,3) {Projection (chunk ${i+1}$)};
\node[rotate around={0:(0,0)},thick, minimum size=3.5ex]  (f0) at (1,1.8) {DR};
\node[rotate around={0:(0,0)},thick, minimum size=3.5ex]  (f0) at (8,1.8) {DR};

\node[rotate around={0:(0,0)},thick, minimum size=3.5ex]  (id0) at (1.5,0.5) {Input (chunk $i$)};
\node[rotate around={0:(0,0)},thick, minimum size=3.5ex]  (id1) at (7.5,0.5) {Input (chunk $i+1$)};
%
\node[minimum size=3.5ex]  (id2) at (4.7,-0.5) {Data stream};
\draw[->,thick,black] (id0)  -- (m0);
\draw[<->,thick,black] (m1)  -- (m0);
\draw[<->,thick,black] (id0)  -- (id1);
\draw[->,thick,black] (id1)  -- (m1);
\draw[-,dotted,thick,black] (-1.9,0.3)  -- (-0.1,0.3);
\draw[-,dotted,thick,black] (9.2,0.3)  -- (10.8,0.3);
\draw[-,thick,black] (0,-0.2)  -- (0,0.2);
\draw[-,thick,black] (3,-0.2)  -- (3,0.2);
\draw[-,thick,black] (6,-0.2)  -- (6,0.2);
\draw[-,thick,black] (9,-0.2)  -- (9,0.2);
\draw[->,thick,black] (-2,0) -- (11.5,0);
%
}}
\caption{Assessing the topological changes in the latent space: comparing the persistent entropy of projected points using a Dimensionality Reduction (DR) technique.}
\label{Descriptor}  
\end{figure}

Recently, it was introduced a clustering method based in SOM for assessing distribution changes in data streams with high dimensional data~\cite{basterrech2024IJCNN}.
SOM is used for projecting the input data into a latent space, then the analysis is done in the latent space, where the authors computed a distance matrix between the input pattern and cluster centers.
The assessment of the distribution shifts is done by applying a statistical summary. This approach of using a data descriptor was also applied in~\cite{Hinder2020,Hinder2023}, and is commonly used in methods based on kernel projections~\cite{Schoplkopf:2002}. 
Here, we modify the framework introduced in~~\cite{basterrech2024IJCNN}, which is based on distances and statistical summaries of the points in the latent space, to an approach that assesses topological changes according to the homological characteristics of the points in the latent space.
Once the DR mapping is done, a distance matrix is computed. 
The distance matrix has the information between the projected point and the cluster centers. 
We use relative locations instead of directly working with the coordinates of points in the latent space generated by the DR method. 
The coordinates are arbitrarily selected and often do not consider any property of the data itself. 
There are even problems in cases where the coordinates are not natural in any sense~\cite{Carlsson2006}. 
Therefore, the relative locations of the point cloud in the latent space are computed by calculating the distances between the mass centers and the projected points.
Hence, our focus is on the geometric properties of the latent space, independent of the chosen coordinates in the latent space.
This methodological approach is illustrated in Figure~\ref{Descriptor}.
Note that the approach is general in the sense that any type of DR technique can be used.

\subsection{Pipeline of the proposed framework}
\label{Pipeline}
%
%

The proposed framework for assessing topological changes based in persistent entropy involves the following steps: 
\begin{itemize}
\item[i)] \textbf{Initial setup}: An initial batch of points is used as a pre-phase procedure to train the selected DR method.

\item[ii)] \textbf{Dimensionality reduction}: When a chunk of data comes, then the input data points are projected using the chosen DR method.

\item[iii)] \textbf{Embedding of the geometrical properties in the latent space}:
For each projected data point, the distance between the projected point and each centroid is computed. By centroid, we refer to the mass center of a cluster, which can be thought of as the average position of all the points within the cluster, or a representative data point that best exemplifies the cluster's characteristics. Then, for each chunk, a matrix is created (with dimensions of number of clusters $\times$ Chunk size) where the column vector has the distances between a projected point and the centroids.

\item[iv)] \textbf{Representation of topological features}: A persistent diagram is created using the distance matrix described in the previous step. The persistence diagram summarizes the information in the distance matrix (note that each projected point is computed a distance matrix). Then, the persistent entropy is computed from the collection of persistence diagrams within the chunk, ignoring the infinity bar~\cite{Persim}. The representation of topological features returns a real vector with a number of elements equal to the size of the chunk.

\item[v)] \textbf{Statistical analysis}. We compute a final index that corresponds to a comparison between the representation of topological features in the current chunk and the representation of topological features in a reference chunk. In our experiments, we compare two consecutive chunks.
%
The comparison between these two vectors is done using a non-parametric test. In our experiments, we applied the Mann-Whitney U test. However, analysis of similarity between these two vectors can be done using other tools. The statistical test provides a p-value score, which we use for monitoring changes between a current chunk of data and a reference chunk of data.
\end{itemize}
Finally, when the global procedure is applied online, it generates a sequence of p-value scores. This sequence provides information about significant changes in the topological properties of the data stream.
Note, an initial training phase is performed to compute the initial clustering and its representative mass centers. 
An initial time window of the data stream is used for training the SOM weights and other global parameters.
After this initial training phase, we continue the learning process following a usual continual learning scenario.
The procedure described previously can include an additional step that involves adjusting the DR technique once a shift is detected. 
When a shift is detected, the SOM method can be re-trained using an arbitrary batch of data (for example, the last chunk or the last $n$ chunks of data), with the precautions to avoid catastrophic forgetting. 
This online calibration procedure was not investigated in this initial empirical exploration and could be a promising direction for future research.
\begin{figure}
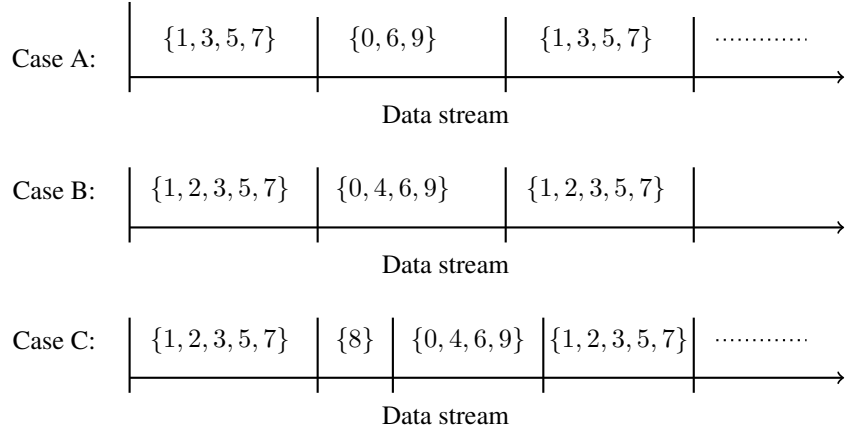

\centering
\scalebox{1}{
\tikz{
\node[rotate around={0:(0,0)},thick, minimum size=3.5ex]  (id1) at (1.2,0.5) {$\{1,3,5,7\}$};
\node[rotate around={0:(0,0)},thick, minimum size=3.5ex]  (id1) at (3.5,0.5) {$\{0,6,9\}$};
\node[rotate around={0:(0,0)},thick, minimum size=3.5ex]  (id1) at (6.2,0.5) {$\{1,3,5,7\}$};
%
\node[thick, minimum size=3.5ex]  (id1) at (4.2,-0.5) {Data stream};
\node[thick, minimum size=3.5ex]  (id1) at (-1,0.25) {Case A:};
\draw[-,black] (7.8,0.5)  edge[dotted,thick] (9,0.5);
\draw[-,black] (0,-0.2)  edge[-,thick] (0,1);
\draw[-,black] (5,-0.2)  edge[-,thick] (5,0.8);
\draw[-,black] (7.5,-0.2)  edge[-,thick] (7.5,0.8);
\draw[-,black] (2.5,-0.2)  edge[-,thick] (2.5,0.8);
\draw[-,black] (0,0) edge[->,thick] (9.5,0);
\node[thick, minimum size=3.5ex]  (id1) at (4.2,-2.5) {Data stream};
\draw[-,black] (0,-2) edge[->,thick] (9.5,-2);
\draw[-,black] (7.5,-2.2)  edge[-,thick] (7.5,-1.2);
\draw[-,black] (2.5,-2.2)  edge[-,thick] (2.5,-1.2);
\draw[-,black] (5,-2.2)  edge[-,thick] (5,-1.2);
\draw[-,black] (0,-2.2)  edge[-,thick] (0,-1.2);
\node[thick, minimum size=3.5ex]  (id1) at (-1,-1.5) {Case B:};
\node[rotate around={0:(0,0)},thick, minimum size=3.5ex]  (id1) at (1.2,-1.5) {$\{1,2,3,5,7\}$};
\node[rotate around={0:(0,0)},thick, minimum size=3.5ex]  (id1) at (3.5,-1.5) {$\{0,4,6,9\}$};
\node[rotate around={0:(0,0)},thick, minimum size=3.5ex]  (id1) at (6.2,-1.5) {$\{1,2,3,5,7\}$};
\node[thick, minimum size=3.5ex]  (id1) at (4.2,-4.5) {Data stream};
\draw[-,black] (0,-4) edge[->,thick] (9.5,-4);
\draw[-,black] (5.5,-4.2)  edge[-,thick] (5.5,-3.2);
\draw[-,black] (2.5,-4.2)  edge[-,thick] (2.5,-3.2);
\draw[-,black] (3.5,-4.2)  edge[-,thick] (3.5,-3.2);
\draw[-,black] (0,-4.2)  edge[-,thick] (0,-3.2);
\draw[-,black] (7.5,-4.2)  edge[-,thick] (7.5,-3.2);
\node[thick, minimum size=3.5ex]  (id1) at (-1,-3.5) {Case C:};
\node[rotate around={0:(0,0)},thick, minimum size=3.5ex]  (id1) at (1.2,-3.5) {$\{1,2,3,5,7\}$};
\node[rotate around={0:(0,0)},thick, minimum size=3.5ex]  (id1) at (3,-3.5) {$\{8\}$};
\node[rotate around={0:(0,0)},thick, minimum size=3.5ex]  (id1) at (4.5,-3.5) {$\{0,4,6,9\}$};
\node[rotate around={0:(0,0)},thick, minimum size=3.5ex]  (id1) at (6.5,-3.5) {$\{1,2,3,5,7\}$};
\draw[-,black] (7.8,-3.5)  edge[dotted,thick] (9,-3.5);
}}
\caption{\label{MNIST-DataStream} Creation of synthetic case studies. Data streams were generated with the MNIST samples interchanged among the different topological types. The graphics illustrate the transition between the sequence of images from one topological type to another type.}
\end{figure}
\section{Experimental results}
In this section, we explore the utility of the proposed approach for monitoring and detecting topological changes in a data stream in the context of continual learning.
We designed the experiments to evaluate and contrast the efficacy of the dimensionality reduction methods previously discussed: PCA, Kernel PCA, and SOM.
%
%
There exists a deficiency in the availability of extensive and diverse real-world data streams within high-dimensional spaces for analyzing the impact of distribution changes~\cite{SouzaChallenges2020}.
This inconvenience is more relevant in the domain of unsupervised analysis of streaming data. 
As a consequence, we created three synthetic data streams with annotated shifts. Our analysis is unsupervised (we do not use the data labels); we use the information about the annotated shifts to evaluate the performance of the proposed approach. 
The created stream has samples from MNIST~\cite{lecun2010}.
The annotations indicate the time stamps where the topological differences between two cloud of points were injected.

\subsection{Benchmark data}
We generated three synthetic datasets using the MNIST dataset~\cite{leCun1989}, following the methodology outlined in~\cite{Basterrech2023DSAA,basterrech2024IJCNN,Rabanser2019}. The procedure consists of creating a stream with chunks of samples that follow a specific distribution, and then alternating these chunks with chunks of instances from a different distribution. 
By construction, the time stamps of the distribution changes are arbitrary predefined. As a consequence, we have marked the exact time stamps at which a drift was injected to evaluate the capacity of the approach in monitoring shifts.
We created the data to check if the method can detect ``essential'' changes between sequences of digits. We divided the digits into three groups: those without any holes (zero-dimensional homology), those with one hole (one-dimensional homology), and those with two holes (one-dimensional homology with two loops).
We denote the three experimental studies as: A, B and C. 
For the three cases, we analyze $20000$ samples. For case studies A and B, the drift is injected every 1000 samples, and for scenario C, the drift occurs every 500 samples.
In case study A, we analyze a data stream where the changes occur between chunks with digits in $\{1,3,5,7\}$ (zero-dimensional homology) and chunks with digits in $\{0,6,9\}$ (one-dimensional homology).
Case study B also includes the digits 2 and 4. These digits are problematic due to variations in handwriting.
Some individuals write the numbers 2 and 4 without any hole, while others write them with one hole, depending on individual handwriting style. 
Then,case study B has a data stream considering the following exchanges between points in $\{1,2,3,5,7\}$ and $\{0,4,6,9\}$.
Finally, in case study C, we evaluate a data stream that includes the number 8, which is not topologically equivalent to any of the other digits.
Case C exchanges samples from the three subsets $\{1,2,3,5,7\}$, $\{0,4,6,9\}$, and $\{8\}$.
Figure~\ref{MNIST-DataStream} depicts how the streaming data was generated for each of the three scenarios. 
Case study A and B have 20 exchanges between subsets containing digits with and without holes, and case study C involves 40 exchanges between subsets containing digits without holes, with one hole, and two holes.
%
%

\subsection{Experimental settings}
For each of the three case studies, we used the first $20\%$ of samples for the initial setup, and we train the parameters of the SOM algorithm.
This training was made offline, as a pre-phase of the continual learning process. 
The SOM algorithm has a grid with $10\times 10$ neurons.
We also analyzed three values for the chunk size parameter $\{50,100,250\}$.
The quality assessment of the monitoring for the shifts was done using a p-value computed with the non-parametric Mann-Whitney U test.
We also evaluate the approach by applying PCA and Kernel-PCA instead of SOM, i.e. we apply PCA and Kernel-PCA in step (ii) of the procedure described in Section~\ref{Pipeline}.
The same initial time window used for SOM was applied to configure PCA and Kernel PCA.
%
Then, we apply the trained DR methods to project the data points into a latent space, following a CL setting.

\subsection{Implementation details}
The implementation of the methods developed during our investigation, as well as the experimental environment, was carried out using the \emph{Python} v3.9 programming language. Several libraries were utilized to facilitate this process, including \emph{NumPy} v1.19.5 for numerical computations, \emph{stream-learn} v0.8.16 for handling data streams, \emph{Sklearn} v1.0.2 for machine learning tasks~\cite{sklearn_api}, and the \emph{Persim} v0.3.6 package for operations related to persistent homology~\cite{Persim,Tralie2018,Bauer2021}.
The source code of our investigation and the datasets are available in the git repository\footnote{\url{https://github.com/sebabaster/Drift-persistence}}.

\subsection{Results}
\begin{figure}[h]
    \centering
\includegraphics[width=\linewidth]{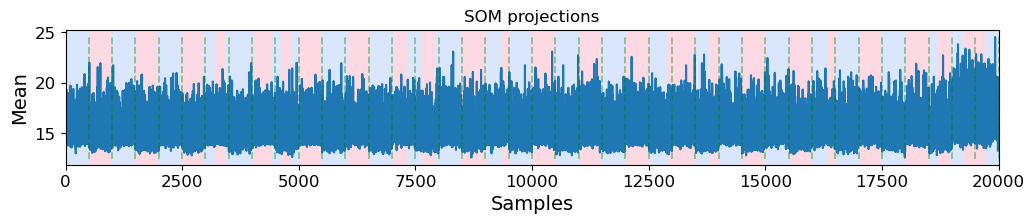}
\caption{Example of the latent space. Off-line analysis of the latent space generated by the SOM projections, and applying change point detection over the distance matrix.}
    \label{SOMPELT}
\end{figure}

\begin{figure}[h]
    \centering
\includegraphics[width=\linewidth]{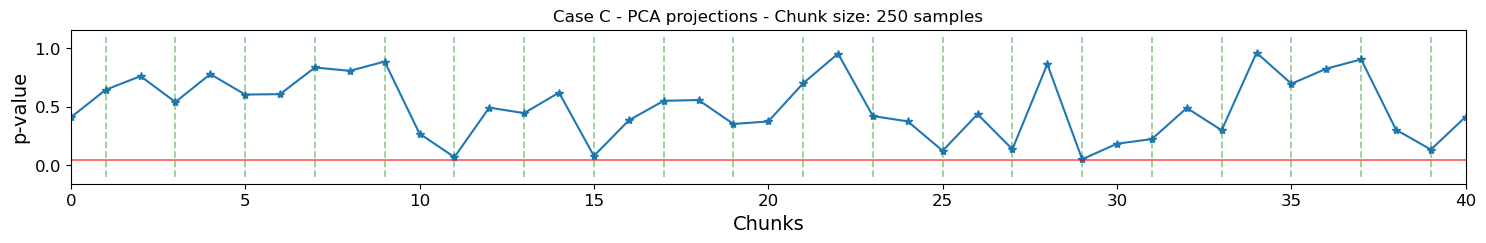}
\includegraphics[width=\linewidth]{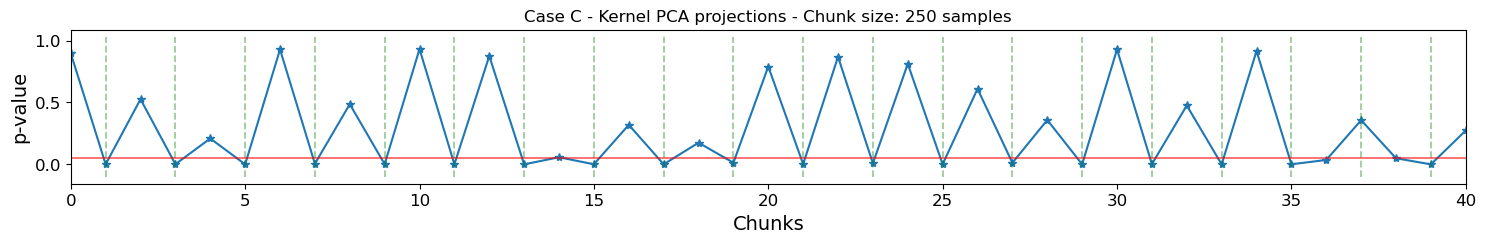}
\includegraphics[width=\linewidth]{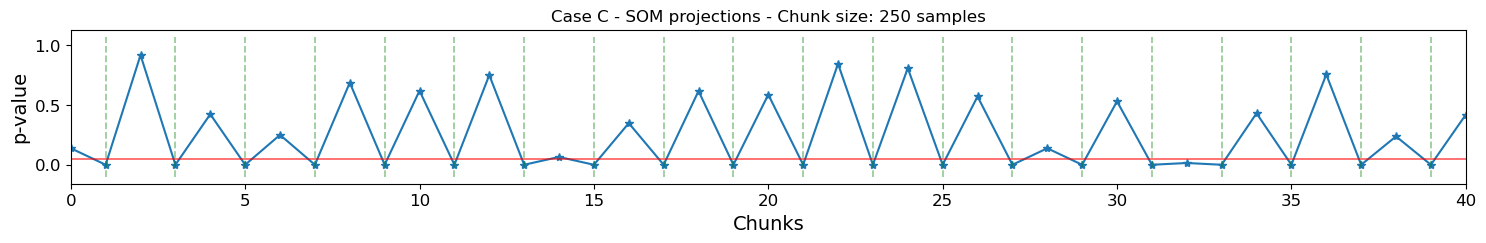}
\caption{Comparison among the three projections. The comparison was made over the dataset C with chunk size of 250 samples.}
\label{CaseC250}
\end{figure}

\begin{figure}[h]
    \centering
\includegraphics[width=\linewidth]{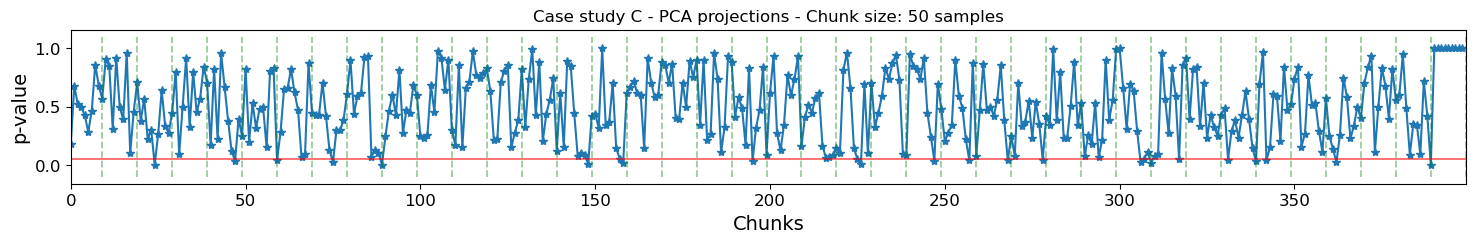}
\includegraphics[width=\linewidth]{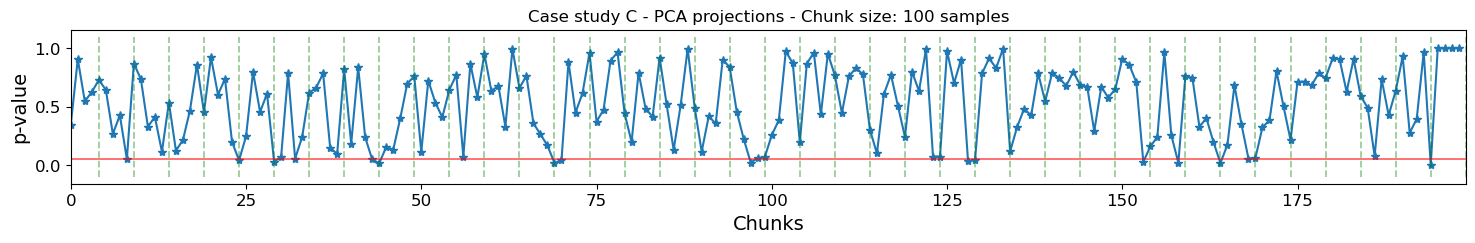}
\caption{Results of the method using PCA projections in case study C for chunk sizes of 50 and 100 samples.}
\label{CaseCSensitivityPCA}
\end{figure}

\begin{figure}[h]
    \centering
\includegraphics[width=\linewidth]{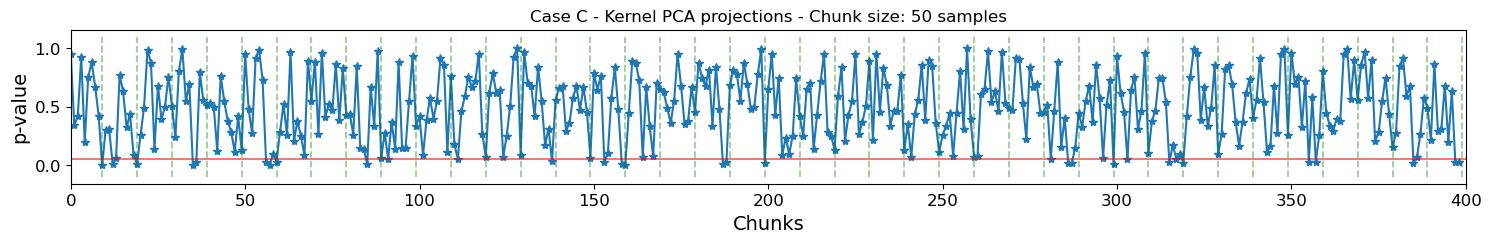}
\includegraphics[width=\linewidth]{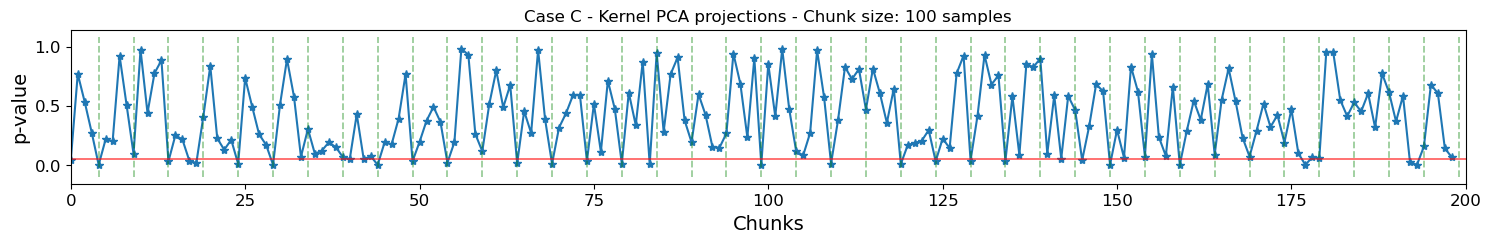}
\caption{Results of the method using Kernel PCA projections in case study C for chunk sizes of 50 and 100 samples.}
\label{CaseCSensitivityKernelPCA}
\end{figure}

\begin{figure}[h]
    \centering
\includegraphics[width=\linewidth]{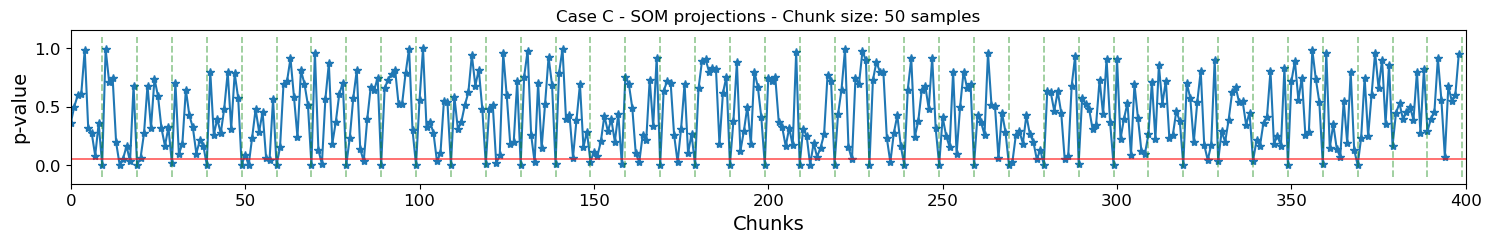}
\includegraphics[width=\linewidth]{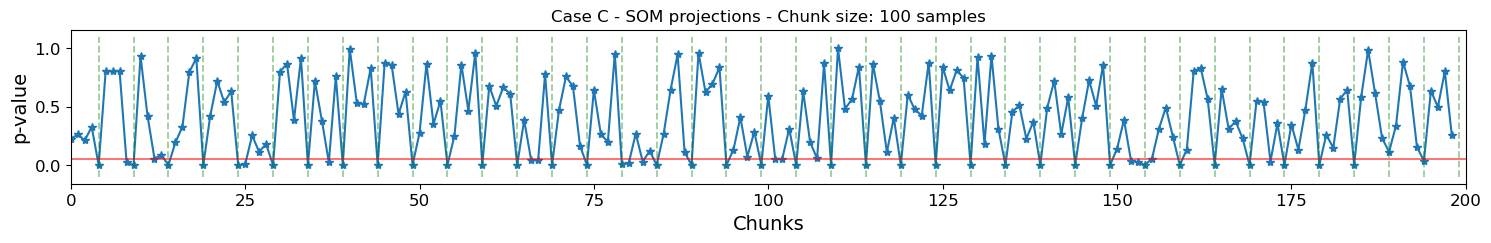}
\caption{Results of the method using SOM projections in case study C for chunk sizes of 50 and 100 samples.}
\label{CaseCSensitivitySOM}
\end{figure}

Figure~\ref{SOMPELT} illustrates the latent space generated by the SOM projections of the data in case C.
The curve (blue dots) in the figure represents the mean distance between the projected points and the mass centers of the clusters. 
It shows the evolution of the mean values for each of these distance matrices. 
In addition, Figure~\ref{SOMPELT} shows the results of applying Pruned Exact Linear Time (PELT) algorithm~\cite{Killick:2012} to detect changes in the sequence of these mean values.
The PELT method is recognized for its computational efficiency, when compared with other change-point detection techniques~\cite{Svoboda2024}.
This experiment was done offline with the purpose of visualizing the complexity of the problem. The aim was to show that detecting shifts in the generated data is not sufficient by merely applying a DR method and computing the mean distance to the mass centers.
We consider Figure~\ref{SOMPELT} as an illustration of the data stream characteristics for case study C. 
This figure also shows the detected change points using the PELT technique. 
The background colors represent the changes detected by PELT, and the vertical green lines indicate the injected shifts. 
The visualization using PELT serves as a representation of the problem’s complexity. It was conducted independently of the other experiments, in which we compare the approaches using SOM, PCA, and Kernel-PCA. 
In other words, we are not comparing PELT to the other techniques; this experiment was conducted to evaluate whether the generated data presents difficulties for a well-known technique for change point detection.  
The other experiments simulate a continual learning environment, making them even more complex than when the problem is addressed offline.
Figure~\ref{CaseC250} presents the results for the three DR techniques. These techniques were evaluated using case study C. The data stream was split into chunks of 250 samples.
The vertical lines (represented by green dashed lines) indicate the injected shifts. The horizontal line represents a p-value of 0.05. According to the results, the linear projection is unable to accurately predict the shifts. This observation is consistent with other studies in the literature that discuss the limitations of linear projections in detecting distribution shifts~\cite{basterrech2024IJCNN,Hinder2022}.
The performance of SOM appears to be slightly better to that of Kernel-PCA.
For instance, refer to the p-values in the chunks between $30$ and $35$.
Chunk size is a crucial parameter for experiments conducted in online settings. The influence of the chunk size is shown in Figures~\ref{CaseCSensitivityPCA},~\ref{CaseCSensitivityKernelPCA}, and~\ref{CaseCSensitivitySOM}, as well as in Table~\ref{tabla}.
Figure~\ref{CaseCSensitivityPCA} shows the results of using linear projections before the analysis of the persistent entropy. There are two graphs: the top graph shows the results for chunks with 50 instances, and the bottom graph shows the results for chunks of 100 samples. This figure also highlights the limitations of linear projections for solving this specific problem, as the method provides few alarms and detects only a low number of drifts.
Figure~\ref{CaseCSensitivityKernelPCA} presents the results of Kernel PCA for chunks with 50 and 100 samples. 
Similar graphics are depicted in Figure~\ref{CaseCSensitivitySOM} where the results of SOM for chunks with 50 and 100 samples are presented.
Table~\ref{tabla} summarizes the results of our experiments, including an evaluation of the impact of chunk size. 
The last two columns show the flags generated by the model using p-values with significance levels of~$0.05$ and~$0.1$. Additionally, we show the number of injected drifts in the experiment. Note that this number is an approximation due to the anomalies that can exist in the datasets (digits with a different number of holes than expected).
According to the table, it seems that SOM may provide results using p-values with a 0.05 level of significance, while Kernel-PCA obtains better results when a p-value with a 0.1 level of significance is considered.
As one might intuitively expect, smaller chunks decrease the quality of geometric pattern analysis. In contrast, larger chunks can encompass more than one shift. 
The optimal chunk size should be determined through experimentation, taking this trade-off into account.

\section{Discussion}
\begin{figure}[h]
    \centering
    \includegraphics[width=0.2\linewidth]{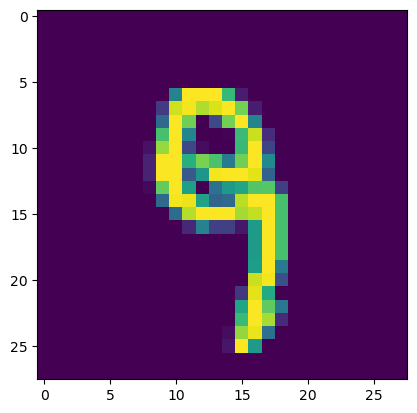}
\includegraphics[width=0.2\linewidth]{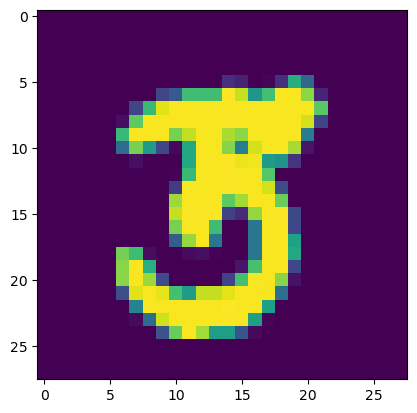}
\includegraphics[width=0.2\linewidth]{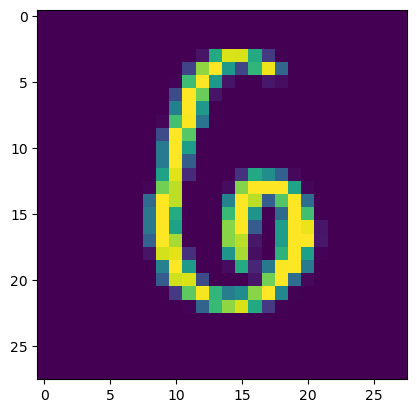}
\includegraphics[width=0.2\linewidth]{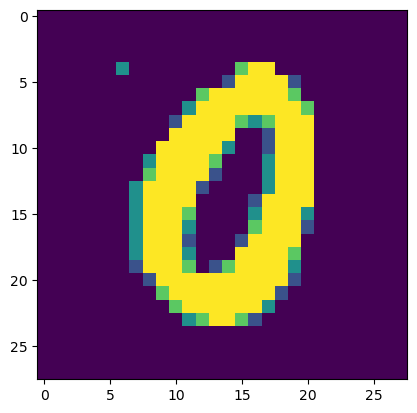}
\caption{An example of digits with a structure different from the one assumed in the case studies. The first image has two holes instead of one. The second image, which can be a number 3 or 5, has a hole. The third image doesn't have any holes. The last image is not a connected component (it has an isolated pixel in the top-left).}
    \label{ExamplesAnomalies}
\end{figure}

As far as we know, this is the first work that attempts to define drifts, including sudden changes in the topological features of the data.
Our research hypothesis was to investigate whether the use of TDA may be helpful or not for monitoring topological changes and detecting ``essential'' differences between chunks of objects.
In other words, if we have a data stream with different types of doughnuts and then suddenly the data contains coffee cups, a ``traditional'' drift detector would detect a drift (using the standard notion of drift~\cite{gama2014survey}). 
However, if we consider purely topological information, both shapes (coffee cups and doughnuts) are considered equivalent. Therefore, it would not be appropriate to consider this type of changes as a drift; in the case that we are interested in detecting ``essential'' differences.
Shapes with the same Euler characteristic cannot be considered as ``essentially'' different.
The inclusion of TDA in the concept drift domain may be highly beneficial for avoiding false alarm when the data has only suffer simple transformations such as scaling, distortions and rotations.
To answer this research question, we implemented a set of experiments based in the MNIST dataset~\cite{leCun1989}. 
%
We defined three case studies based on changes in the Euler characteristic of the data.
%
The digits were not individually visually inspected; instead, they were sampled using a uniform distribution. Consequently, there may be instances where an image was expected to have a specific number of holes, but the sample shows a different number. We illustrate these anomaly examples in Figure~\ref{ExamplesAnomalies}. The figure contains four graphics. The first image has at least two holes, whereas the expected number is one. The second image has one hole, but it represents either the number three or five, both of which typically have zero holes. The third image represents the number six, but it was not properly drawn, so the circle is not closed. Finally, we also show an image with pixels that are not connected, forming more than one connected component.
Therefore, the analyzed scenarios contain noisy marks in the stream.
%
For further analyses, it would be beneficial to generate new datasets with higher quality annotated data streams.
In this study, we did not evaluate different strategies to mitigate the catastrophic forgetting problem. When a drift is detected, it is necessary to define an appropriate strategy to fine-tune and update the framework. This is especially important for SOM. We left this evaluation for future research.
Another limitation of our experimental analysis is that the framework is composed of two modules: dimensionality reduction and persistent entropy. Further investigation is required to analyze the relevance of each of these modules and how they affect drift detection.
Additionally, an ablation study considering the different steps in the pipeline would enhance the reliability of our experiments.

Last but not least, defining concept drift solely based on changes in topological features appears to present an incomplete understanding of the phenomenon.
Analyzing objects only based on their topological aspects ignores the meaning of the objects. 
Human interpretation, including  contextual insights and semantic understanding of the information, may also be relevant in some problems.
For example, 6 and 9 are different digits according to semantic understanding, but a simple rotation can transform one digit into the other.
Therefore, to define drift ignoring human interpretation, which is the case of purely consider topology also have limitations.
Combining the three aspects--statistics and geometry (both considered in previous literature), and topology (the approach introduced here)--in the definition of drifts could be a promising direction.
An additional interesting angle would be to also integrate the meaning of the information.
Therefore, there are several avenues for future exploration, presenting many open questions that remain to be addressed.

\section{Conclusions and future work}
We introduced a novel approach to concept drift detection, leveraging algebraic topology and persistent entropy. We broaden the scope of concept drift, which is typically associated solely with statistical distribution changes in incoming data. In the explored approach, we also integrate the definition of drift to detect significant changes in topological features. 
The framework uses the SOM algorithm to transform the input space, reducing its dimensionality while preserving its topological characteristics. Then, the analysis of data drifts is conducted in the latent space. We explore the potential of persistent entropy to identify significant differences among data from consecutive chunks. We showed the performance of the method over three case studies (based on the  MNIST dataset), and we compared the performance with PCA and Kernel-PCA. The proposed method does not make any assumption about the data distribution. Additionally, it can be applied to both supervised and unsupervised problems.  
We believe that this work is an initial step towards applying TDA in the area of concept drift. A potential direction for future research could involve evaluating the framework with different types of data streams, including dynamic graphs. Furthermore, it would be interesting to compare persistent entropy with other measures.
\begin{table}[]
        \caption{Sensitive analysis of the chunk size for the three study cases, the three methods, and two levels of significance of the hypothesis test (0.05 and 0.1). The table shows the number of potential drifts (flags) detected by each method. The fourth column shows the number of injected drifts. This is an approximate number because, during the construction of the dataset, each image wasn't individually verified. Consequently, digits like 2, 4, 6, and 9 may sometimes have a hole and sometimes not.}
    \centering
\scalebox{0.9}{
    \begin{tabular}{c|c|c|c|c|c}
\hline
\hline
    Case study & Chunk size & Method &  Drifts &  \makecell{Flags \\ (p-value at $0.05$ level)} &  \makecell{Flags \\(p-value at $0.1$ level)}\\ 
    \hline
\multirow{9}{*}{A} & \multirow{3}{*}{50} & SOM & \multirow{3}{*}{20} & 13 & 30 \\
 &  & PCA &  & 22 & 51 \\
 &  & KernelPCA &  & 27 & 58 \\ \cline{2-6}
 & \multirow{3}{*}{100} & SOM & \multirow{3}{*}{20} & 9 & 16 \\
 &  & PCA &  & 10 & 27 \\
 &  & KernelPCA &  & 14 & 22 \\ \cline{2-6}
 & \multirow{3}{*}{250} & SOM & \multirow{3}{*}{20} & 3 & 5 \\
 &  & PCA &  & 6 & 11 \\
 &  & KernelPCA &  & 4 & 8 \\ \hline \cline{2-6}
\multirow{9}{*}{B} & \multirow{3}{*}{50} & SOM & \multirow{3}{*}{20} & 18 & 36 \\
&  & PCA &  & 17 & 34 \\
 &  & KernelPCA &  & 17 & 39 \\ \cline{2-6}
 & \multirow{3}{*}{100} & SOM & \multirow{3}{*}{20} & 10 & 26 \\
 &  & PCA &  & 6 & 12 \\
 &  & KernelPCA &  & 10 & 21 \\ \cline{2-6}
 & \multirow{3}{*}{250} & SOM & \multirow{3}{*}{20} & 4 & 12 \\
 &  & PCA &  & 3 & 4 \\
 &  & KernelPCA &  & 9 & 14 \\ \hline 
\multirow{9}{*}{C} & \multirow{3}{*}{50} & SOM & \multirow{3}{*}{40} & 53 & 75 \\
 &  & PCA &  & 21 & 44 \\
 &  & KernelPCA &  & 27 & 51 \\\cline{2-6}
 & \multirow{3}{*}{100} & SOM & \multirow{3}{*}{40} & 43 & 53 \\
 &  & PCA &  & 13 & 26 \\
 &  & KernelPCA &  & 26 & 37 \\\cline{2-6}
 & \multirow{3}{*}{250} & SOM & \multirow{3}{*}{40} & 40 & 43 \\
 &  & PCA &  & 8 & 10 \\
 &  & KernelPCA &  & 25 & 35 \\
\hline
\hline
    \end{tabular}
    }
    \label{tabla}
\end{table}
\clearpage
%
%
%
\bibliographystyle{unsrt}
\bibliography{References}

\end{document}